# Dynamic Probabilistic Network Based Human Action Recognition

Anne Veenendaal[1], Eddie Jones[1], Zhao Gang[1], Elliot Daly[1], Sumalini Vartak[2], Rahul Patwardhan[3]

[1]*Computer Science and Emerging Research Lab, AU*
[2]*Lead IT, USA*
[3]*Infobahn Softworld Inc, USA*

## Abstract

This paper examines use of dynamic probabilistic networks (DPN) for human action recognition. The actions of lifting objects and walking in the room, sitting in the room and neutral standing pose were used for testing the classification. The research used the dynamic interrelation between various different regions of interest (ROI) on the human body (face, body, arms, legs) and the time series based events related to the these ROIs. This dynamic links are then used to recognize the human behavioral aspects in the scene. First a model is developed to identify the human activities in an indoor scene and this model is dependent on the key features and interlinks between the various dynamic events using DPNs. The sub ROI are classified with DPN to associate the combined interlink with a specific human activity. The recognition accuracy performance between indoor (controlled lighting conditions) is compared with the outdoor lighting conditions. The accuracy in outdoor scenes was lower than the controlled environment.

## Introduction

Probabilistic networks are very useful to establish the causal relationship between two inter-related events. Such networks are suitable for linking dependencies between variables in the problem domain which could vary from speech recognition, vision based systems, medicine or automotive industries. The dynamic probabilistic networks need a large amount of training data to be effective. Co-incidentally, the video image frames contain an abundance of dynamic motion based images and movement of features which can be exploited for training the network. Several studies [1], [2], [3], [4] to [10] in the last decade have concentration on using the temporal aspects of features from video and image sequences for face recognition, gesture recognition and human activity. As a result this study examines the use of DPN to detect specific actions inside a controlled environment such as walking, sitting, lifting objects and also tested the performance of the system under outdoor lighting to measure the generalizability of the recognition system. Moreover the focus is shifting from static image analysis to research on temporal changes in series of images for detecting human activities. The performance of recognition systems under low lighting or outdoor conditions and in case of noisy background and chaotic and uncontrolled conditions is still poor. Human activity recognition is also closely related to more complex tasks such as emotion recognition. Several studies [11], [12], [13], [14] to [28] have examined the multimodal emotion recognition techniques and the software implementation strategies. Studies[29], [30], [31], [32] to [44] have used joint probability distributions, modelling of activities using manifold and learning their transition probabilities. Hidden Markov model (HMM) based recognition has also gained interest. These methods are practically infeasible and difficult to implement because of the computationally intensive nature of model training process and dependency on the availability of large amount of training data. Additionally there is a higher degree of co-relation between the features in the data and the network making the accuracy confined and over-fitted to the intra-corpus feature properties. Another strategy currently being explored is the exemplar based approach such as view specific manifolds and spatio-temporal embedded classification using statistical learning and Bayesian inference techniques. But these methods have the same interdependency between the statistical model and the exemplar classes.

## Method

The video of 6 participants walking in a room and lifting objects was recorded. The resolution of the video was set to very low to minimize the image frame size. Each image frame was set to a size of 640 x 480 pixel window. The video of participants resulted in 24 action scenes. Each video consisted of 1500 to 2000 frames and 5-6 minute videos. First each action sequence was manually identified and annotated to establish the ground truth. For the purpose of training the DBN, the data was split into 10 small sets containing a sequence of lifting an object and keeping it down randomly chosen from the master list. Feature extraction was performed on each sequence and the feature vector was used to train the DBN. For the feature vector a bounding box was used with the corner co-ordinates of the box, the average intensity value of the change in pixel intensities for pixels within the box was calculated and the filling ratio and first order moments around the centroid were used. These features formed a feature vector with 8 components. The feature vectors were set as the observational input for the DBN training. The priors and transition matrices defaults were chosen with random automated event detection.

The various components of the human body were detected using cascade transform to identify the region of interest (ROI). Once the bounding box was detected, the features were extracted using projection methods followed by principle component analysis (PCA). After the dimensionality reduction the bounding box was used to obtain features for the face, body, hands and legs (4 components). Thus the feature vector consisted of 8 x 4 = 32 features representing an activity ranging from walking, sitting, lifting objects and standing in a neutral position. According to the model for Bayesian inference, the detection of activity from an image sequence is done by the maximum a posteriori (MAP) decision rule of probability distribution. This probability is depends on the prior probability of each class and the likelihood of image sequences given each class. Additionally, the value is normalized overall the observations such that sum of probabilities over all classes is 1. To reduce the computational processing of the model training, every third image sequence frame was considered from a frame rate of 24 per second as a result only 8 frames were considered. The prediction of human activity recognition was implemented using the majority vote based classification scheme. The prediction of each class for the 4 components was also calculated separately and the classifiers for each component (face, hand, body, leg) were used for voting for a class. In case of a tie, the probability of hand was used because of the more dynamic nature of the feature movement while performing an activity and the lack of course facial features from a long distance between the camera and the subject. The performance of the classifier for the various human activity classes is discussed in the results section.

## Result

The recognition accuracy was better in case of non-noisy video sequences. For the noisy sequences the recognition accuracy dropped from 78% to 74% for picking up action and dropped from 79% to 71% for keeping object down on the floor.

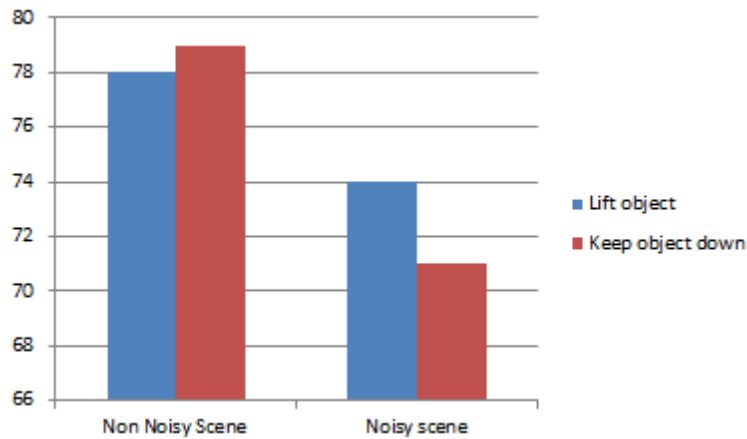

Fig. 1. DBN recognition accuracy

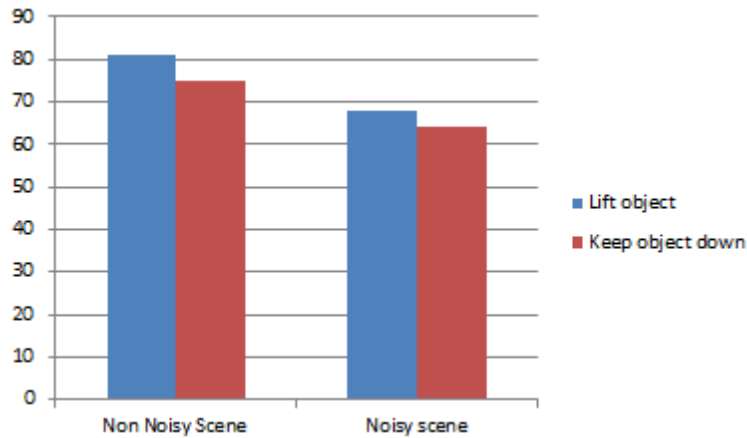

Fig. 2. HMM recognition accuracy

A comparison with HMM based human activity recognition indicated that HMM was recognition rate was better for lifting action but dropped by a higher degree in case of noisy scene as compared to DBN technique.

The results from the test video sequence for the dynamic probabilistic network are shown below and it was obtained by the MAP decision rule maximization. The plot of posterior probability Fig. 3 of each class given the sequence of test images showed that the DPN was capable of accurate recognition of various activities such as walking, sitting, lifting objects compared under controlled conditions. A comparison of recognition rates for the various activities under indoor and outdoor conditions using majority voting scheme is also shown in Fig. 4. In each instance the recognition rates of indoor controlled conditions was better.

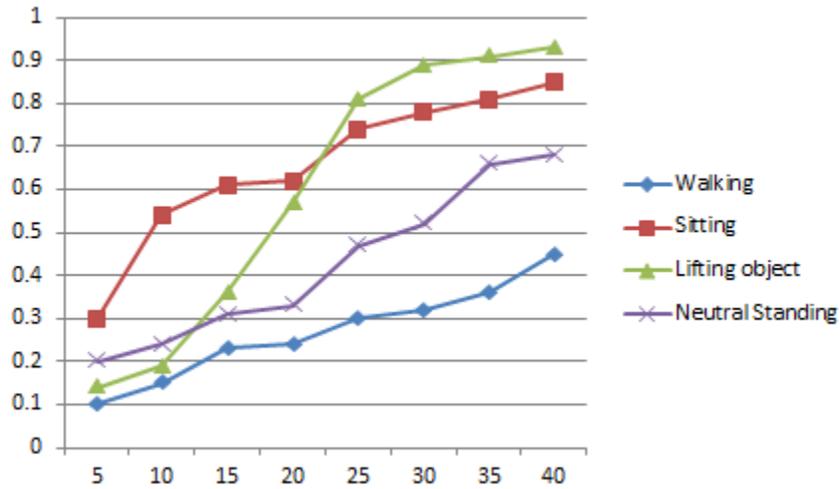

Fig. 3. Comparison of posterior probabilities for various

The probability of lifting objects steadily improved as more data was available from subsequent image sequences. The remaining actions probabilities steadily improved before flattening out indicating that the subsequent video frames could not provide discriminating information about the activity.

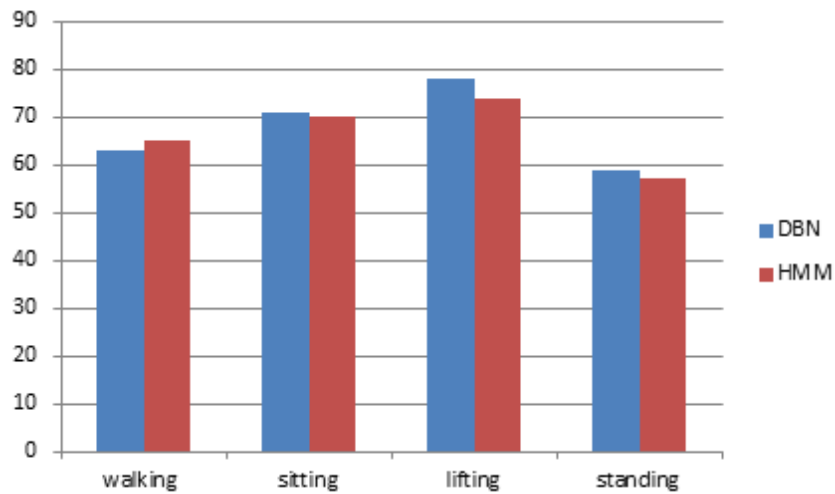

Fig. 4. Comparison of accuracies using DBN and HMM.

The DBN performed better for the lifting object activity under controlled lighting, standing, and sitting. In case of the activities representing walking, HMM had a better accuracy over the proposed method.

## Conclusion

The DBN based human activity recognition was more robust compared to HMM. The proposed method for using region of interest and then combining the features from face, hand, body and legs to form a feature vector proved useful for recognition under controlled conditions and reducing the computational complexity. But the same strategy was not useful in outdoor scenes. This was evident from the fact that the recognition rate dropped by a smaller value as compared to that of HMM in case of noisy scene. Additionally, the DBN based technique was more suitable for busy video scenes with several temporal changes and activity as compared to only individual actors in the recorded scene and the corresponding HMM based recognition.